\def\year{2018}\relax
\documentclass[letterpaper]{article} 
\usepackage{aaai18}  
\usepackage{times}  
\usepackage{helvet}  
\usepackage{courier}  
\usepackage{url}  
\usepackage{graphicx}  
\usepackage[font=small,labelfont=bf]{caption}

\usepackage{matharticle}
\usepackage{paralist}
\usepackage{todonotes}
\usepackage{xspace}
\newcommand{\horse}{\texttt{Lipizzaner}\xspace}
\newcommand{\horsed}{\texttt{Lipizzaner-D}\xspace}

\usepackage[utf8]{inputenc} 
\usepackage[T1]{fontenc}    
\usepackage{hyperref}       
\usepackage{url}            
\usepackage{booktabs}       
\usepackage{amsfonts}       
\usepackage{nicefrac}       
\usepackage{microtype}      
\usepackage{comment}
\usepackage[numbers]{natbib}

\algblock{ParFor}{EndParFor}
\algnewcommand\algorithmicparfor{\textbf{parfor}}
\algnewcommand\algorithmicpardo{\textbf{do}}
\algnewcommand\algorithmicendparfor{\textbf{end\ parfor}}
\algrenewtext{ParFor}[1]{\algorithmicparfor\ #1\ \algorithmicpardo}
\algrenewtext{EndParFor}{\algorithmicendparfor}

\algrenewcommand\alglinenumber[1]{\tiny #1:}

\begin{document}
%
\title{Towards Distributed Coevolutionary GANs}
\author{Abdullah Al-Dujaili, Tom Schmiedlechner, Erik Hemberg and Una-May O’Reilly\\
aldujail@mit.edu, tschmied@mit.edu, hembergerik@csail.mit.edu, unamay@csail.mit.edu\\
 CSAIL, MIT, USA\\
}
\maketitle

\begin{abstract}
	Generative Adversarial Networks (GANs) have become one of the dominant methods for deep generative modeling. Despite their demonstrated success on multiple vision tasks, GANs are difficult to train and much research has been dedicated towards understanding and improving their gradient-based learning dynamics.
	Here, we investigate the use of coevolution, a class of black-box (gradient-free) co-optimization techniques and a powerful tool in evolutionary computing, as a supplement to gradient-based GAN training techniques. Experiments on a simple model that exhibits several of the GAN gradient-based dynamics (e.g., mode collapse, oscillatory behavior, and vanishing gradients) show that coevolution is a promising framework for escaping degenerate GAN training behaviors. 
\end{abstract}

\section{Introduction}
\label{sec:intro}

Generative modeling aims to learn functions that express distributional outputs. In a standard setup, generative models take a training set drawn from a specific distribution and learn to represent an estimate of that distribution. By estimate, we mean either an explicit density estimation, the ability to generate samples, or the ability to do both~\cite{goodfellow2016nips}. GANs~\citep{goodfellow2014generative} are a framework for training generative deep models via an adversarial process. They have been applied with celebrated success to a growing body of applications. 
Typically, a GAN pairs two networks, viz. a generator and a discriminator. The goal of the generator is to produce a sample (e.g., an image) from a latent code such that the distribution of the produced samples are indistinguishable from the true data (training set) distribution. In tandem, the discriminator plays the role of a critic to do the assessment and tell whether the samples are true data or generated by the generator. Concurrently, the discriminator is trained to discriminate optimally (maximize its accuracy), while the generator is trained to fool the discriminator (minimize its accuracy). Despite their witnessed success, it is well known that GANs are difficult to optimize. From a game theory perspective, GAN training can be seen as a two-player minimax game. Since the two networks are differentiable, optimizing the minimax GAN objective is typically carried out by (variants of) simultaneous gradient-based updates to their parameters. While it has been shown that simultaneous gradient updates converge if they are made in \emph{function space}, the same proof does not apply to these updates in parameter space~\cite{goodfellow2014generative}. On the one hand, a zero gradient is a necessary condition for standard optimization to converge. On the other hand, \emph{equilibrium} is the corresponding necessary condition in a two-player game~\cite{arora2017generalization}. In practice, gradient-based GAN training often oscillates without ultimately reaching an equilibrium. Moreover, a variety of degenerate behaviors have been observed---e.g., \emph{mode collapse}~\cite{arora2017gans}, \emph{discriminator collapse}~\cite{li2017towards},  and \emph{vanishing gradients}~\cite{arjovsky2017towards}. These unstable learning dynamics have been the focus of several investigations by the deep learning community, seeking a principled theoretical understanding as well as practical algorithmic improvements and heuristics~\cite{arjovsky2017towards,arjovsky2017wasserstein,gulrajani2017improved}. 
Two-player minimax black-box optimization and games have been a topic of recurrent interest in the evolutionary computing community~\cite{
laskari2002particle,qiu2017new}.  In seminal work, Hillis~\cite{hillis1990coev} showed that more efficient \emph{sorting programs} can be produced by competitively co-evolving them versus their \emph{testing programs}. Likewise, Herrmann~\cite{herrmann1999genetic} proposed a two-space genetic algorithm as a general technique to solve minimax optimization problems and used it to solve a parallel machine scheduling problem with uncertain processing times. In \emph{competitive coevolution}, two different populations, namely solutions and tests, coevolve against each other~\cite{floreano2008bio}. The quality of
a solution is determined by its performance when interacting with the tests. Reciprocally, a test's quality is determined by its performance when interacting with the solutions, leading to what is commonly referred to as evolutionary arms race~\cite{dawkins1979arms}.

In this paper, we propose to pair a coevolutionary algorithm with conventional GAN training, asking whether the combination is powerful enough to more frequently avoid degenerate training behaviors. The motivation behind our proposition is of two-fold. First, most of the pathological behaviors encountered with gradient-based GAN training have been identified and studied by the evolutionary computing community decades ago---e.g., \emph{focusing}, \emph{relativism}, and \emph{loss of gradients}~\cite{nolfi1998coevolving,watson2001coevolutionary}. Second, there has been a growing body of work, which shows that the performance of gradient-based methods can be rivaled by evolutionary-based counterparts when combined with sufficient computing resources and data~\cite{salimans2017evolution,morse2016simple,lehman2017more,clune2017uber}. The aim of this paper is to bridge the gap between works of the deep learning and evolutionary computing communities towards a better understanding of gradient-based and gradient-free GAN dynamics. Indeed, one can see that the Nash Equilibrium solution concept in coevolutionary literature~\cite{popovici2012coevolutionary} is not that different from the notion of GAN mixtures in GAN literature~\cite{arora2017generalization}.

We report the following contributions: \begin{inparaenum}[\itshape i)]
	\item For a simple parametric generative modeling problem~\cite{li2017towards} that exhibits several degenerate behaviors with gradient-based training, we validate the effectiveness of combining coevolution with gradient-based updates (mutations).
	\item We present \horse
	, a coevolutionary framework to train GANs with gradient-based mutations (for neural net parameters) and gradient-free mutations (for hyperparameters) and learn a mixture of GANs.
\end{inparaenum}


\section{Related Work}
\label{sec:background}

\paragraph{Training GANs} Several gradient-based GAN training variants have been proposed to improve and stabilize its dynamics. One variant category is focused on improving training techniques for single-generator single-discriminator networks. Examples include modifying the generator's objective~\cite{warde2016improving}, the discriminator's objective~\cite{metz2016unrolled}, or both~\cite{arjovsky2017wasserstein,salimans2016improved}. Some of these propositions are theoretically well-founded, but convergence still remains elusive in practice. The second category employs a framework of multiple generators and/or multiple discriminators. Examples include training multiple discriminators~\cite{durugkar2016generative}; training an array of specialized discriminators, each of which looks at a different random low-dimensional projection of the data~\cite{neyshabur2017stabilizing}; sequentially training and adding new generators with boosting techniques~\cite{tolstikhin2017adagan}; training a cascade of GANs~\cite{wang2016ensembles}; training multiple generators and discriminators in parallel (GAP)~\cite{jiwoong2016generative}; training a classifier, a discriminator, and a set of generators~\cite{hoang2017multi}; and optimizing a weighted average reward over pairs of generators and discriminators (MIX+GAN)~\cite{arora2017generalization}. 
For a theoretical view on GAN training, the reader may refer to~\cite{arjovsky2017towards,li2017towards,arora2017generalization}.

\paragraph{Coevolutionary Algorithms for Minimax Problems.} Variants of competitive coevolutionary algorithms have been used to solve minimax formulations in the domains of constrained optimization~\cite{barbosa1999coevolutionary}, mechanical structure optimization~\cite{Barbosa1997ACG}, and machine scheduling~\cite{herrmann1999genetic}. These early coevolutionary propositions were tailored to symmetric minimax problems. In practice, the symmetry property may not always hold. In fact, mode collapse in GANs may arise from asymmetry~\cite{goodfellow2016nips}. To address this issue, \emph{asymmetric fitness evaluation} was presented in~\cite{jensen2003new} and analyzed in \cite{branke2008new}. Further, Qiu et al.~\cite{qiu2017new} attempt to overcome the limitations of existing coevolutionary approaches in solving minimax optimization problems using differential evolution. 



\section{Methods}
\label{sec:methods}

\paragraph{Notation}
We adopt a mix of notation used in \cite{arora2017generalization, li2017towards}.
Let $\calG=\{G_u, u \in \calU\}$ denote the class of generators, where $G_u$ is a function indexed by $u$ that denotes the parameters of the generators. Likewise, let $\calD=\{D_v, v \in \calV\}$ denote the class of discriminators, where $D_v$ is a function parameterized by $v$. Here $\calU, \calV \subseteq \R^{p}$ represent the parameters space of the generators and discriminators, respectively.  Further, let $G_*$ be the target unknown distribution that we would like to fit our generative model to. Formally, the goal of GAN training is to find parameters $u$ and $v$ so as to optimize the objective function
\begin{equation}
\footnotesize
\label{eq:gan-def}
\min_{u \in \calU}\max_{v \in \calV} \calL(u,v)\;, \;\text{where} \nonumber
\end{equation}
\begin{equation}
\footnotesize
\calL(u,v) = \E_{x\sim G_*}[\phi (D_v(x))] + \E_{x\sim G_u}[\phi(1-D_v(x))]\;,
\end{equation}
and $\phi:[0,1] \to \R$, is a concave function, commonly referred to as the \emph{measuring function}. In the recently proposed Wasserstein GAN~\cite{arjovsky2017wasserstein}, $\phi(x)=x$, and we use the same for the rest of the paper. In practice, we have access to a finite number of training samples $x_1, \ldots, x_S \sim G_*$. Therefore, one can use the empirical version $\frac{1}{S}\sum_{i=1}^{S} \phi(D_v(x_i))$ to estimate  $\E_{x\sim G_*}[\phi (D_v(x))]$ . The same holds for $G_u$. Further, let $\calS_u$ be a distribution supported on $\calU$ and $\calS_v$ be a distribution supported on $\calV$.

\paragraph{Basic Coevolutionary Dynamics.} With coevolutionary algorithms, the two search spaces $\calU$ and $\calV$ can be searched with two different populations: the generator population $P_u=\{u_1, \ldots, u_T\}$ and the discriminator population $P_v=\{v_1, \ldots, v_T\}$, where $T$ is the population size. In a predator-prey interaction, the two populations coevolve: the generator population $P_u$ aims to find generators which evaluate to low $\calL$ values with the discriminator population $P_v$ whose goal is to find discriminators which evaluate to high $\calL$ values with the generator population. This is realized by harnessing the neo-Darwanian notions of heredity and survival of the fittest, as outlined in Algorithm~\ref{alg:basic-coev-gan}. Over multiple generations (iterations), the fitness of each generator $u_i\in P_u$ and discriminator $v_j \in P_v$ are \emph{evaluated} based on their interactions with one or more discriminators from $P_v$ and  generators from $P_u$, respectively (Lines~\ref{line:evl-beg} to~\ref{line:evl-end}). Based on their fitness \emph{rank} (Lines~\ref{line:sel-beg} to~\ref{line:sel-end}), the current population individuals are employed in producing  next population of generators and discriminators with the help of \emph{mutation}: a genetic-like variation operator (Lines~\ref{line:mut-beg} to~\ref{line:mut-end}), where the mutated individuals replace the current population if they exhibit a better fitness. In gradient-free scenarios, Gaussian mutations are usually applied~\cite{qiu2017new,reckless2018ash}. With GANs (which are differentiable nets), we propose to use gradient-based mutations for the generators and discriminators net parameters, i.e., $P_u$ and $P_v$ are mutated with a gradient step computed by back-propagating through one (or more) of their fitness updates (right-hand side of Lines~\ref{line:fit-update-beg} and~\ref{line:fit-update-end}). Note that the coevolutionary dynamics are not restricted to tuning net parameters. Non-differentiable (hyper)parameters can also be incorporated. In our framework, we tune the learning rates for the generator and discriminator populations with Gaussian mutations.

\begin{algorithm}[h!]
	\floatname{algorithm}{\tiny Algorithm}
	\tiny
	\caption{\tiny  \texttt{BasicCoevGANs}($P_u, P_v, \calL, \{\alpha_{i}\}, \{\beta_{i}\},I$) \newline
		\textbf{Input:} \newline  
		~$P_u$~: generator population \hspace{11em} 
		~$P_u$~: discriminator population \newline 
		~$\{\alpha_{i}\}$~: selection probability\hspace{10em}
		~$\{\beta_{i}\}$~: mutation probability \newline 
		~$I$~: number of generations \hspace{11em}
		~$\calL$~: GAN objective function \newline 
		\textbf{Return:} \newline 
		~$P_u$~: evolved generator population \newline 
		~$P_v$~: evolved discriminator population \newline 
	}
	\label{alg:basic-coev-gan}
	\begin{algorithmic}[1]
		 \bf
		\For {$i$ in $range(I)$}
		\Statex \hspace{1.5em}// Evaluate $P_u$ and $P_v$
		\State $f_{u_1\ldots u_T} \gets 0$ \label{line:evl-beg}
		\State $f_{v_1\ldots v_T} \gets 0$
		\For {each $u_i$ in $P_u$, each $v_j$ in $P_v$}
		\State $f_{u_i} \mathrel{-}= \calL(u_i, v_j)$ \label{line:fit-update-beg}
		\State $f_{v_j} \mathrel{+}= \calL(u_i, v_j)$ \label{line:fit-update-end}
		\EndFor  \label{line:evl-end}
		\Statex \hspace{1.5em}// Sort $P_u$ and $P_v$
		\State $u_{1\ldots T} \gets u_{s(1)\ldots s(T)} \texttt{ with } s(i)=\texttt{argsort} (f_{u_1\ldots u_T} , i)$ \label{line:sel-beg}
		\State $v_{1\ldots T} \gets v_{s(1)\ldots s(T)} \texttt{ with } s(j)=\texttt{argsort} (f_{v_1\ldots v_T} , j)$
		\Statex \hspace{1.5em}// Selection 
		\State $u_{1\ldots T} \gets u_{s(1)\ldots s(T)}  \texttt{ with } s(i)=\texttt{argselect} (u_{1\ldots T}  , i, \{\alpha_i\})$
		\State $v_{1\ldots T}  \gets v_{s(1)\ldots s(T)} \texttt{ with } s(j)=\texttt{argselect} (v_{1\ldots T}, j, \{\alpha_{j}\})$ \label{line:sel-end}
		\Statex \hspace{1.5em}// Mutation \& Replacement
		\State $u_{1\ldots T}\gets \texttt{replace}(\{u_{i}\},\{u^\prime_{i}\}) \texttt{ with } u^\prime_i=\texttt{mutate} (u_i, \beta_i)$ \label{line:mut-beg}
		\State $v_{1\ldots T}  \gets \texttt{replace}(\{v_{j}\},\{v^\prime_{j}\}) \texttt{ with } v^\prime_j=\texttt{mutate} (v_j, \beta_j)$ \label{line:mut-end}
		\EndFor
		\State \Return $P_u, P_v$ \label{line:basic-coev-return}
	\end{algorithmic}

\end{algorithm}

\paragraph{Spatial Coevolution Dynamics.} The basic coevolutionary setup (as adapted for GAN training in Algorithm~\ref{alg:basic-coev-gan}) has been the subject of several studies~(e.g., \cite{watson2001coevolutionary,mitchell2006coevolutionary}) analyzing degenerate behaviors such as focusing, relativism, and loss of gradients; which correspond to mode collapse, discriminator collapse, and vanishing gradients in the GAN literature, respectively. Consequently, this has led to the emergence of more stable setups such as \emph{spatial coevolution}, where individuals from both populations are distributed spatially (e.g., on a grid), with local interactions governing fitness evaluation, selection, and mutation. This is different from the basic coevolutionary setup in which individuals from the two populations test each other either exhaustively or employ random sampling to realize interactions~\cite{mitchell2006coevolutionary}. Spatial coevolution has shown to be substantially successful over several non-trivial learning tasks due to its ability to maintain diversity in the population for long periods and to foster continuing arms races. We refer the reader to \cite{williams2005investigating,mitchell2006coevolutionary} for detailed numerical experiments on the efficiency of spatial coevolution. In the context of GAN training, we distribute the generator and discriminator populations over a two-dimensional toroidal grid where each cell holds one (or more) individual(s) from the generator population and one (or more) individual(s) from the discriminator population. During the coevolutionary process, each cell (and the individuals therein) interacts with its neighboring cells. A cell's \emph{neighborhood} is defined by its adjacent
cells and specified by its size $s_{n}$. A five-cell neighborhood (one
center and four adjacent cells) is a commonly used setup. Note that for an $m\times m$-grid, there exist $m^2$ neighborhoods. 
For the $k$th neighborhood in the grid, we refer to the set of generator individuals in its center cell by $P^{k,1}_u\subset P_u$ and the set of generator individuals in the rest of the neighborhood cells by $P^{k,2}_u, \ldots, P^{k,s_n}$, respectively. Furthermore, we denote the union of these sets by $P^k_u = \cup^{s_n}_{i=1} P^{k,i}_u \subseteq P_u$, which represents the $k$th generator neighborhood. Note that, with $s_n=5$ and for the $k^\prime$th neighborhood  whose center cell's generator individuals  $P^{k^\prime,1}_u = P^{k,j}_u$ for some $j \in \{2,\ldots, s_n\}$, we have $P^k_u \cap P^{k^\prime}_u= P^{k,1}_u \cup P^{k^\prime,1}_u$. Furthermore, $|P^k_u|=|P^{k^\prime}_u|$ for all $k, k^\prime \in \{1,\ldots,m^2 \}$ and we denote this number by $N$. The same notation and terminology is adopted for the discriminator population, with $P^k_v \subseteq P_v$ representing the $k$th discriminator neighborhood. As shown in Algorithm~\ref{alg:adv-coev-gan}, each neighborhood $k$ runs an instance of Algorithm~\ref{alg:basic-coev-gan} with the generator and discriminator populations being $P^k_u$ and $P^k_v$, respectively. The difference is that the evolved populations~(Line~\ref{line:basic-coev-return} of Algorithm~\ref{alg:basic-coev-gan}) are used to update \emph{only} the individuals of the center cells $P^{k,1}_u$, $P^{k,1}_v$ rather than $P^{k}_u$, $P^{k}_v$ (Lines~\ref{line:topn-1} and~\ref{line:topn-2} of Algorithm~\ref{alg:adv-coev-gan}). Since there are $m^2$ neighborhoods, all of the populations individuals will get updated as  $P_u = \cup^{m^2}_{k=1}P^{k}_u$, $P_v=\cup^{m^2}_{k=1}P^{k}_v$. The $m^2$ instances of Algorithm~\ref{alg:basic-coev-gan} can run in parallel in a synchronous or asynchronous fashion (in terms of reading/writing to the populations). In our implementation, we opted for the asynchronous mode for three reasons. First, asynchronous variant scales more efficiently with lower communication overhead among cells. Second, with asynchronous mode, different
cells are often in different stages of the training process (i.e., compute different
generations). Individuals from previous or upcoming generations may therefore
be used during the training process, which further increases the diversity as well~\cite{nolfi1998coevolving, popovici2012coevolutionary}. Third, several works have concluded that asynchronous coevolutionary computing produces slightly better results with less function evaluations~\cite{nielsen2012novel}. 
\paragraph{Generator Neighborhood As A Generator Mixture.} Towards the end of training, $|P_u|$ generators will be available for use as generative models. And instead of using one, we propose to choose one of the generator neighborhoods $\{P^k_u\}_{1\leq k \leq m^2}$ as a mixture of generators according to a given performance metric $g:\calU^{N} \times \R^{N}  \to \R$ (e.g., inception score~\cite{salimans2016improved}). That is, the best generator mixture $P^*_u\in \calU^{N} $ and the corresponding mixture weights $\vw^* \in [0,1]^N$---{Recall that in a neighborhood, there are $N$ generators (and $N$ discriminators). Hence, the $N$-dimensional mixture weight vector $\vw$.}---is defined as follows
\begin{equation}
\footnotesize
P^*_u, \vw^* =\argmax_{P^{k}_u, \vw^k: 1\leq k \leq m^2  } g\big(\sum_{{u_i \in P^{k}_u \\ w_i \in \vw^k}} w_i G_{u_i}\big)\;,
\label{eq:mixture}
\end{equation}
where $w_i$ represents the mixture weight of (or the probability that a data point comes from) the $i$th generator in the neighborhood, with $\sum_{w_i \in \vw^k} w_i = 1$. One may think of $\{\vw^k\}_{1\leq k \leq m^2}$ as hyperparameters of the proposed framework that can be set a priori (e.g., uniform mixture weights $w_i=\frac{1}{N}$). Nevertheless, the system is flexible enough to incorporate learning these weights in tandem with the coevolutionary dynamics as discussed next.

\begin{figure}[t!]
	\centering
	\includegraphics[width=0.35\textwidth]{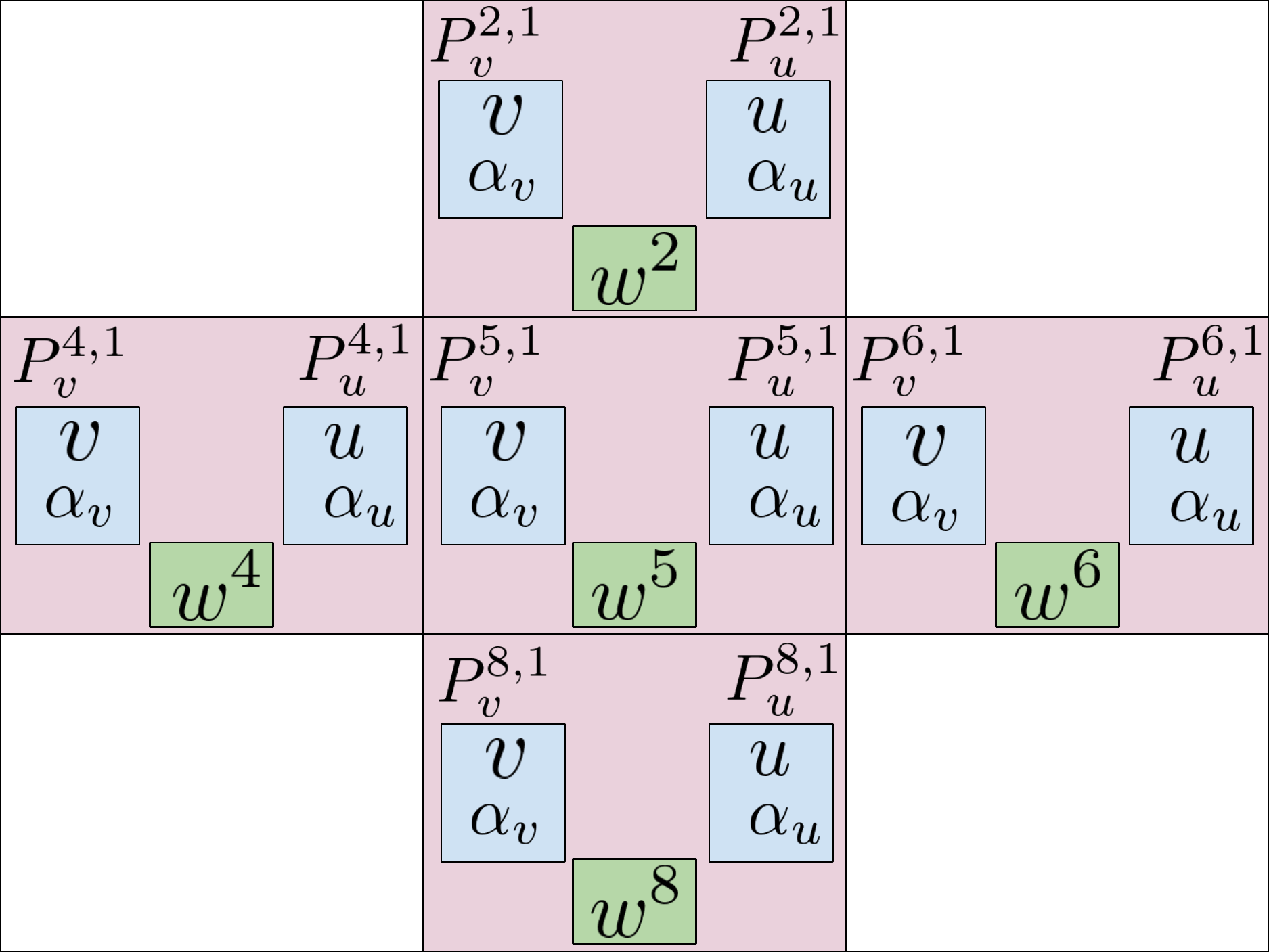}
	\caption{\small   Topology of a $3\times 3$-grid ($m=3$) with a neighborhood
		size of $s_n=5$. A neighborhood of the $5$th cell is highlighted in light red. Each cell has a population size of one (one generator $G_u$
		and one discriminator $D_v$). The corresponding neural net parameters $u$ and $v$ are updated with gradient-based mutations, while the respective hyperparameters (e.g., learning rate
		$\alpha_{u}$ and $\alpha_{v}$) are updated with Gaussian-based mutations based on the interactions of each cell with its neighbors. Each cell has the mixture weight vector
		$\vw^k$ for their respective neighborhood, which is optimized with an evolutionary algorithm according to a given performance metric $g$. 
	}
	\label{fig:spatial-grid}
\end{figure}

\paragraph{Evolving Mixture Weights.} With an $m\times m$-grid, we have $m^2$ mixture weight vectors $\{\vw^k\}_{1\leq k \leq m^2}$ , which we would like to learn and optimize such that our performance metric $g$ is maximized across all the $m^2$ generator neighborhoods. To this end, we view  $\{\vw^k\}_{1\leq k \leq m^2}$ as a population of $m^2$ individuals whose fitness measures are evaluated by $g$ given the corresponding generator neighborhoods. In other words, the fitness of the $k$th individual  (weight vector $\vw^k$) is $g\big(\sum_{\substack{u_i \in P^{k}_u, w_i \in \vw^k}} w_i G_{u_i}\big)$. After each step of spatial coevolution of the generator and discriminator populations, the mixture weight vectors $\{\vw^k\}_{1\leq k \leq m^2}$ are updated with an evolution strategy (e.g., (1+1)-ES \cite[Algorithm 2.1]{loshchilov2013surrogate}), where selection and mutation based on the neighborhoods' $g$ values (Line~\ref{line:es} of Algorithm~\ref{alg:adv-coev-gan}). This concludes the description of our coevolutionary proposition for training GANs with gradient-based mutations as summarized in Algorithm~\ref{alg:adv-coev-gan}. Fig.~\ref{fig:spatial-grid} provides a pictorial illustration of the grid. We refer to our python implementation of Algorithm~\ref{alg:adv-coev-gan} by \horse. 
\begin{algorithm}[h!]
	\floatname{algorithm}{\tiny Algorithm}
	\tiny 
	\caption{
		\tiny 
		\texttt{CoevGANs}($P_u, P_v, \calL, \{\alpha_{i}\}, \{\beta_{i}\}$) \newline
		\textbf{Input:} \newline  
		~$P_u$~: generator population \hspace{12em}
		~$P_u$~: discriminator population \newline 
		~$\{\alpha_{i}\}$~: selection probability\hspace{11em}
		~$\{\beta_{i}\}$~: mutation probability \newline 
		~$I$~: number of population generations per training step \hfill
		~$m$~: side length of the spatial square grid \newline 
		~$\calL$~: GAN objective function \newline
		\textbf{Return:} \newline 
		~$P^*_u$~: evolved generator mixture \hspace{10em}
		~$\vw^*$~: evolved mixture weight vector \newline 
	}
	\label{alg:adv-coev-gan}
	\begin{algorithmic}[1]
		\bf
		\Repeat
		\Statex \hspace{1.5em}// Spatial Coevolution of Generator \& Discriminator Populations
		\ParFor{$k$ in $range(m^2)$} \label{line:parfor-beg}
		\State $\hat{P}^k_u, \hat{P}^k_v \gets$ \texttt{BasicCoevGANs}($P^k_u, P^k_v, \calL, \{\alpha_{i}\}, \{\beta_{i}\},I$)
		\State ${P}^{k,1}_u\gets \texttt{TopN}(\hat{P}^k_u,n=|P^{k,1}_u|$) \label{line:topn-1}
		\State ${P}^{k,1}_v\gets \texttt{TopN}(\hat{P}^k_v,n=|P^{k,1}_v|$) \label{line:topn-2}
		\EndParFor \label{line:parfor-end}
		\Statex \hspace{1.5em}// Generator Mixture Weights Evolution
		\State $\vw^1, \ldots, \vw^{m^2}\gets$\texttt{(1+1)-ES}($\vw^1, \ldots, \vw^{m^2}, g, \{P^k_u)\}$) \Comment{See \cite[Algorithm 2.1]{loshchilov2013surrogate})} \label{line:es}
		\Until training converged
		\State $P^*_u, \vw^* \gets \argmax_{P^{k}_u, \vw^k: 1\leq k \leq m^2  } g\big(\sum_{\substack{u_i \in P^{k}_u \\ w_i \in \vw^k}} w_i G_{u_i}\big)$
		\State \Return $P^*_u, \vw^*$ \label{line:adv-coev-return}
	\end{algorithmic}
\end{algorithm}

\section{Experiments}
\label{sec:experiments}

Two different types of experiments were conducted: \begin{inparaenum}[\itshape 1)]
	\item To elaborate the capability of coevolutionary algorithms to solve typical problems of GANs, we used the theoretical model proposed in~\cite{li2017towards} that exhibits degenerate training behavior in a typical
	framework and compare them when trained with a simple coevolutionary counterpart.
	\item We then show the ability of Lipizzaner to match state-of-the-art GANs on commonly used image-generation datasets \cite{arora2017generalization, arjovsky2017wasserstein}.
\end{inparaenum}

\subsection{Theoretical GAN Model}

\paragraph{Setup.}To investigate coevolutionary dynamics for GAN training, we make use of the simple problem introduced in~\cite{li2017towards}. Formally, the generator set is defined as
\begin{equation}
\footnotesize
\label{eq:gens}
\calG = \bigg\{ \frac{1}{2}\normal(\mu_1,1) + \frac{1}{2} \normal(\mu_2, 1)\mid \vmu \in \R^2 \bigg\} \; .
\end{equation}
On the other hand, the discriminator set is expressed as follows.
\begin{equation}
\footnotesize
\label{eq:discs}
\calD = \{\I_{[\ell_1, r_1]} + \I_{[\ell_2, r_2]} \mid \vell, \vr \in\R^2 ~\mbox{s.t.}~\ell_1 \leq r_1 \leq \ell_2 \leq r_2 \} \; .
\end{equation}
Given a true distribution $G_{*}$ with parameters $\vmu^*$, the GAN objective of this simple problem can be written as
\begin{equation}
\footnotesize
\label{eq:mu-g}
\min_{\vmu} \max_{\vell, \vr} \calL(\vmu, \vell, \vr)\;, \; where \nonumber
\end{equation}
\begin{equation}
\footnotesize
\label{eq:loss}
\calL(\vmu, \vell, \vr) = \E_{x\sim G_{*}}[D_{\vell,\vr}(x)] + \E_{x\sim G_{\vmu}}[1 - D_{\vell, \vr}(x)]\;.
\end{equation}

While being simple to understand and demonstrate, this GAN variant exhibits the relevant dynamics we are elaborating. We conducted several experiments to understand the performance of the coevolutionary framework in its simplest form in comparison to the standard gradient-based dynamic. Unless stated otherwise, we used Algorithm~\ref{alg:basic-coev-gan} with 120 runs per experiment, each run is set with 100 generations and a population size of 10. We also use Gaussian mutation with a step size of 1 as the only genetic operator.

\begin{figure}[t!]
	\centering
	\resizebox{0.5\textwidth}{!}{
		\begin{tabular}{cc}
			\includegraphics[width=0.24\textwidth]{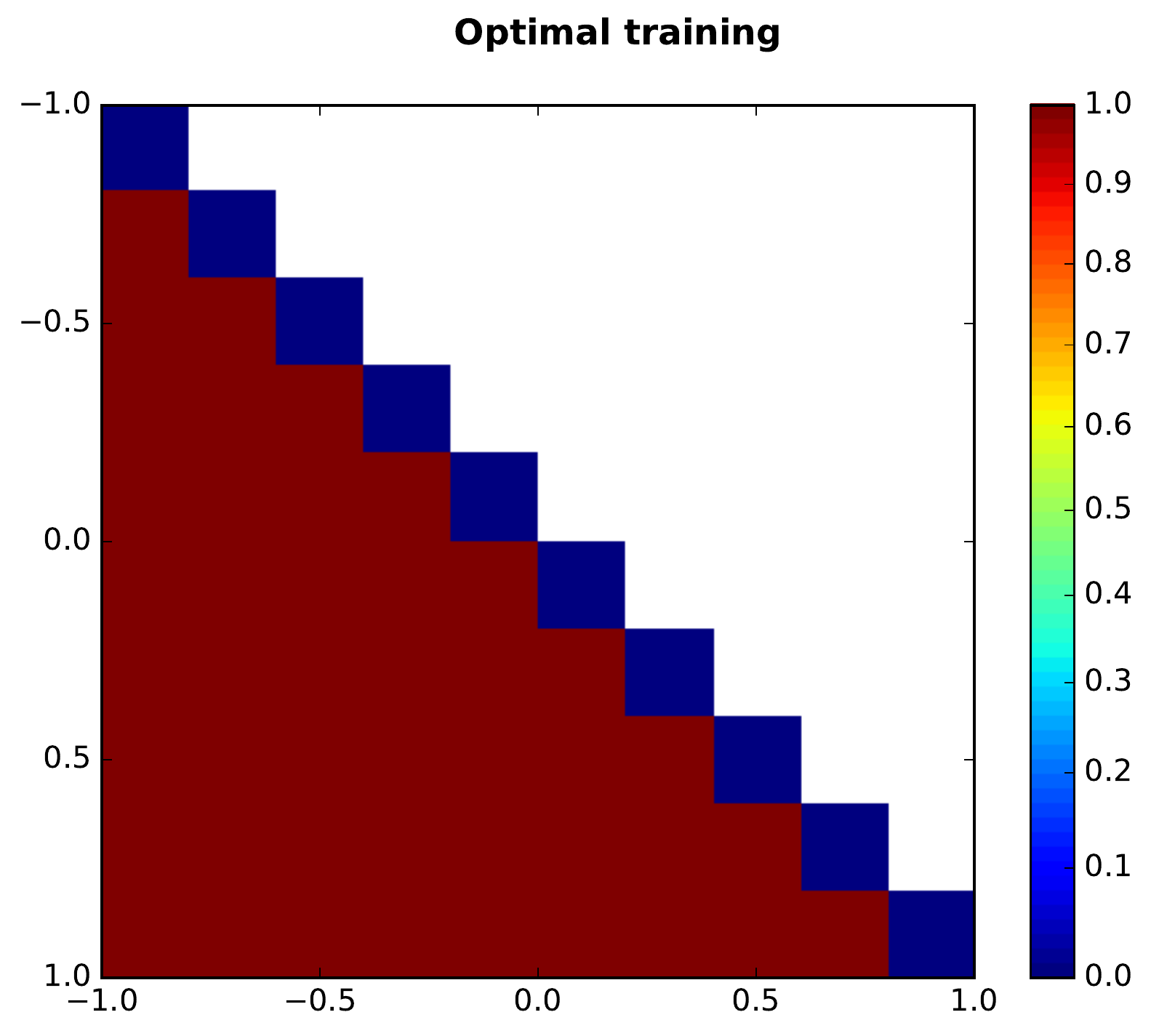} &
			\includegraphics[width=0.31\textwidth,trim=0cm 0.7cm 0cm 0cm, clip]{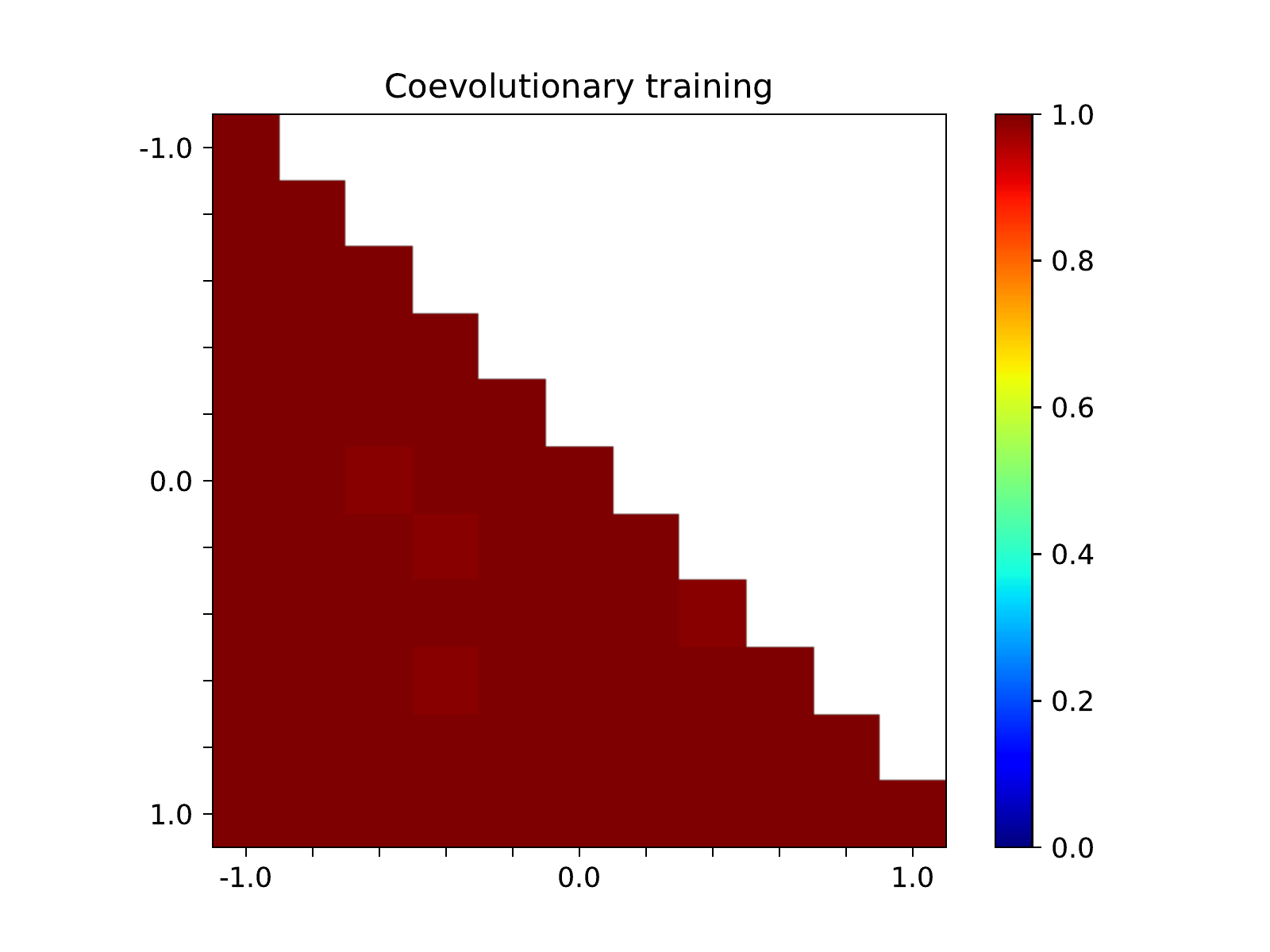} \\
			\bf {\Large (a)} & \bf {\Large (b)}
	\end{tabular}}	
	\caption{\small   Heatmap of success probability for random \textbf{generator} ($\mu_1$ and $\mu_2$) initializations for (a) a variant of gradient-based dynamics (adapted from~\cite{li2017towards}) and (b) coevolutionary GAN training dynamics. For each square, the individuals of the generator population are initialized within the corresponding range.}
	\label{fig:mode-collapse}
\end{figure}

\paragraph{Results.} Fig.~\ref{fig:gan-convergence} shows the convergence of the parameters $\ell_1,\ell_2, r_1, r_2, \mu_1, \mu_2$ using different variants of gradient-based and coevolutionary dynamics. One can observe that the $\mu_1$ and $\mu_2$  under coevolutionary dynamics consistently converge to the true values $\mu^*_1$ and $\mu^*_2$, respectively. Furthermore, we investigated coevolutionary behavior for the following scenarios that have been shown to be critical for traditional pure gradient-based GAN training methods \cite{arora2017gans, li2017towards}: 

\emph{Mode collapse.} Being one of the most-observed failures of GANs in real-world problems, mode collapse often occurs when attempting to learn models from highly complex distributions, e.g. images of high visual quality \cite{arora2017gans}. In this scenario, the generator is unable to learn the full underlying distribution of the data, and attempts to fool the discriminator by producing samples from a small part of this distribution. Vice versa, the discriminator learns to distinguish real and fake values by focusing on another part of the distribution -- which leads to the generator specializing on this area, and furthermore to oscillating behavior. In our experiments, we used the same setting as Li et al. \cite{li2017towards}, initializing $\mu_1$ and $\mu_1$ to values in the interval of $[-10, 10]$, with a step size of $0.1$. Fig.~\ref{fig:mode-collapse} shows the average success rate with the given initialization values. In accordance with \cite{li2017towards}, we define \emph{success} as the ability to reach a distance less than $0.1$, between the best generator of the last generation and the optimal generator $G_*$. From the figure, we see that coevolutionary GAN training is able to step out of mode collapse scenarios, where $\mu_1=\mu_2$---Note the high success rate along the diagonal of Fig.~\ref{fig:mode-collapse}~(b) in comparison to best of gradient-based dynamics in (a).

\begin{figure}[t!]
	\centering
	\resizebox{0.5\textwidth}{!}{
		\begin{tabular}{ccc} {
			\includegraphics[width=0.6\textwidth,clip, trim=0.95cm 0cm 0.95cm 2cm]{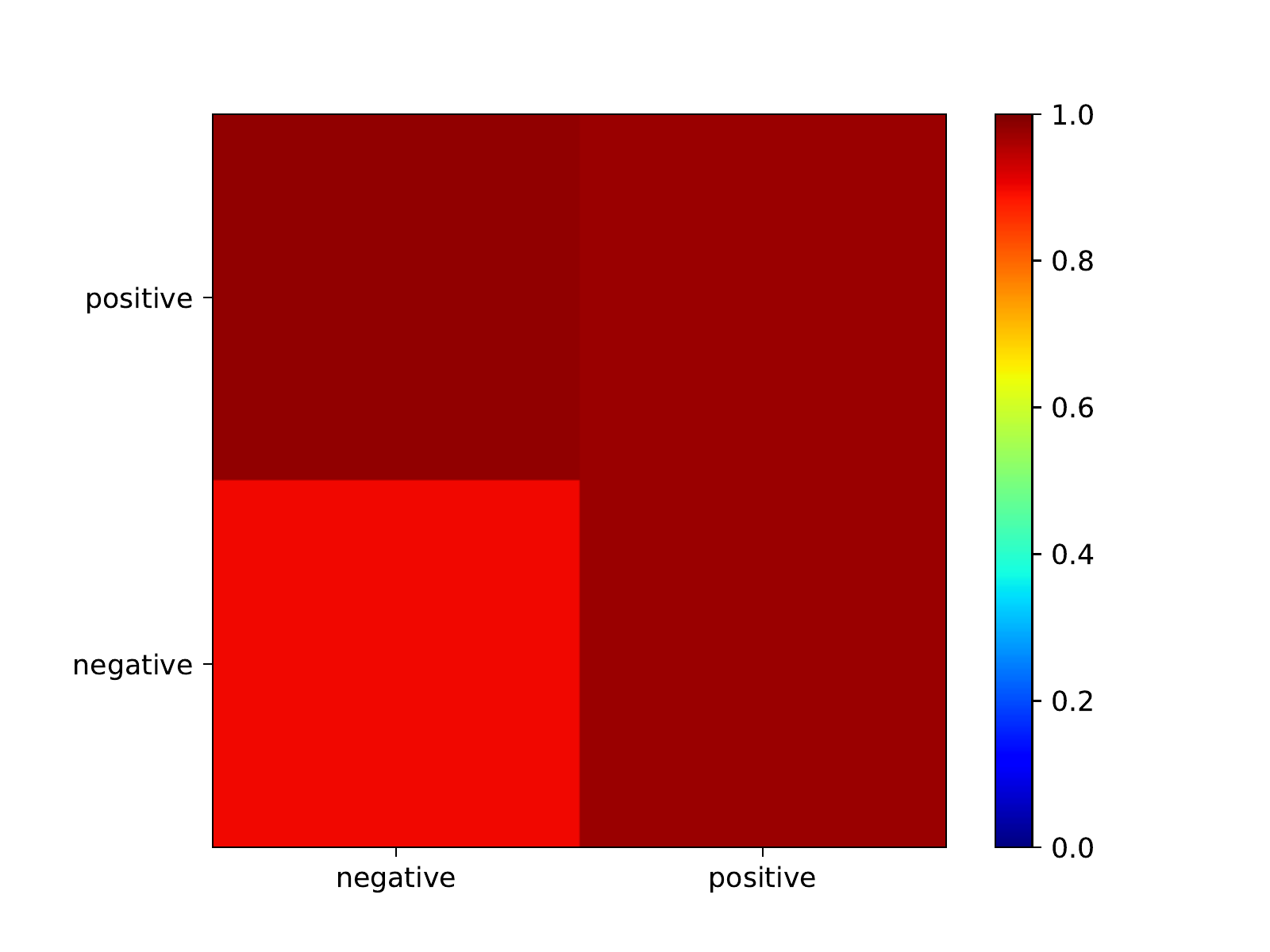} } &
			\includegraphics[width=0.6\textwidth,clip, trim=0.25cm 0cm 0.95cm 0.9cm]{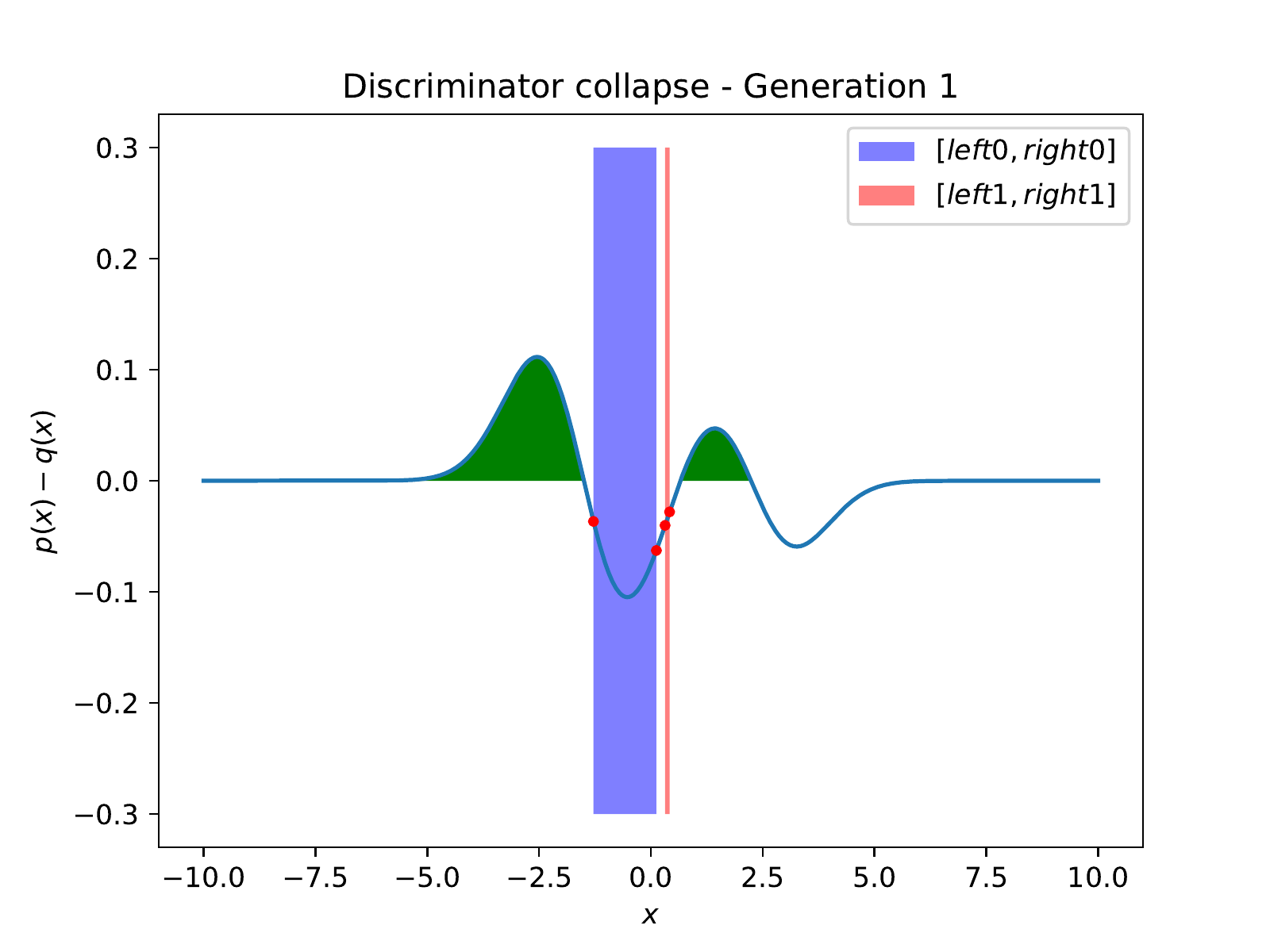}  & 
			\includegraphics[width=0.6\textwidth,clip, trim=0.25cm 0cm 0.95cm 0.9cm]{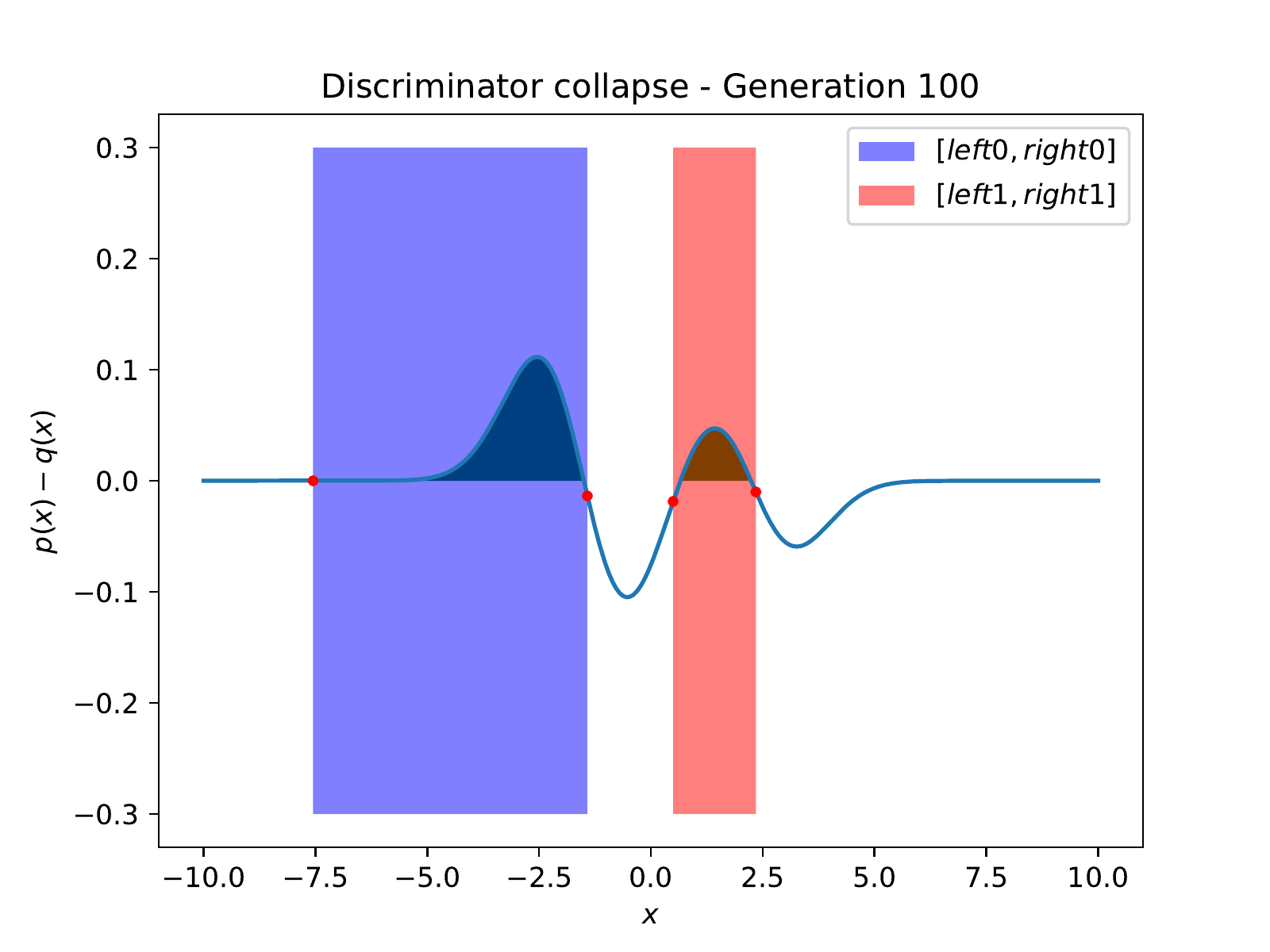}\\
			\bf \Huge (a) & \bf \Huge (b) & \bf \Huge (c) \\
		\end{tabular}	
	}
	\caption{\small  (a) Heatmap of success probability for random \textbf{discriminator}  ($\ell_1,r_1, \ell_2,r_2$) initializations for coevolutionary GAN training dynamics. The axes refer to initial fitness values for both the left ($[\ell_1,r_1]$) and the right ($[\ell_2,r_2]$) bounds, leading to four different quadrants. (b) First and (c) last generation of a coevolutionary trained GAN, initialized in a discriminator collapse setup, which corresponds to the bottom left quadrant of (a). In other words, for each square, the individuals of the discriminator population are initialized randomly such that the signs of their fitness values match those of the corresponding square.}
	\label{fig:discriminator-collapse-progress}
\end{figure}

\begin{figure}[t!]
	\centering
	\resizebox{0.47\textwidth}{!}{
		\begin{tabular}{cc}
			\includegraphics[width=0.35\textwidth]{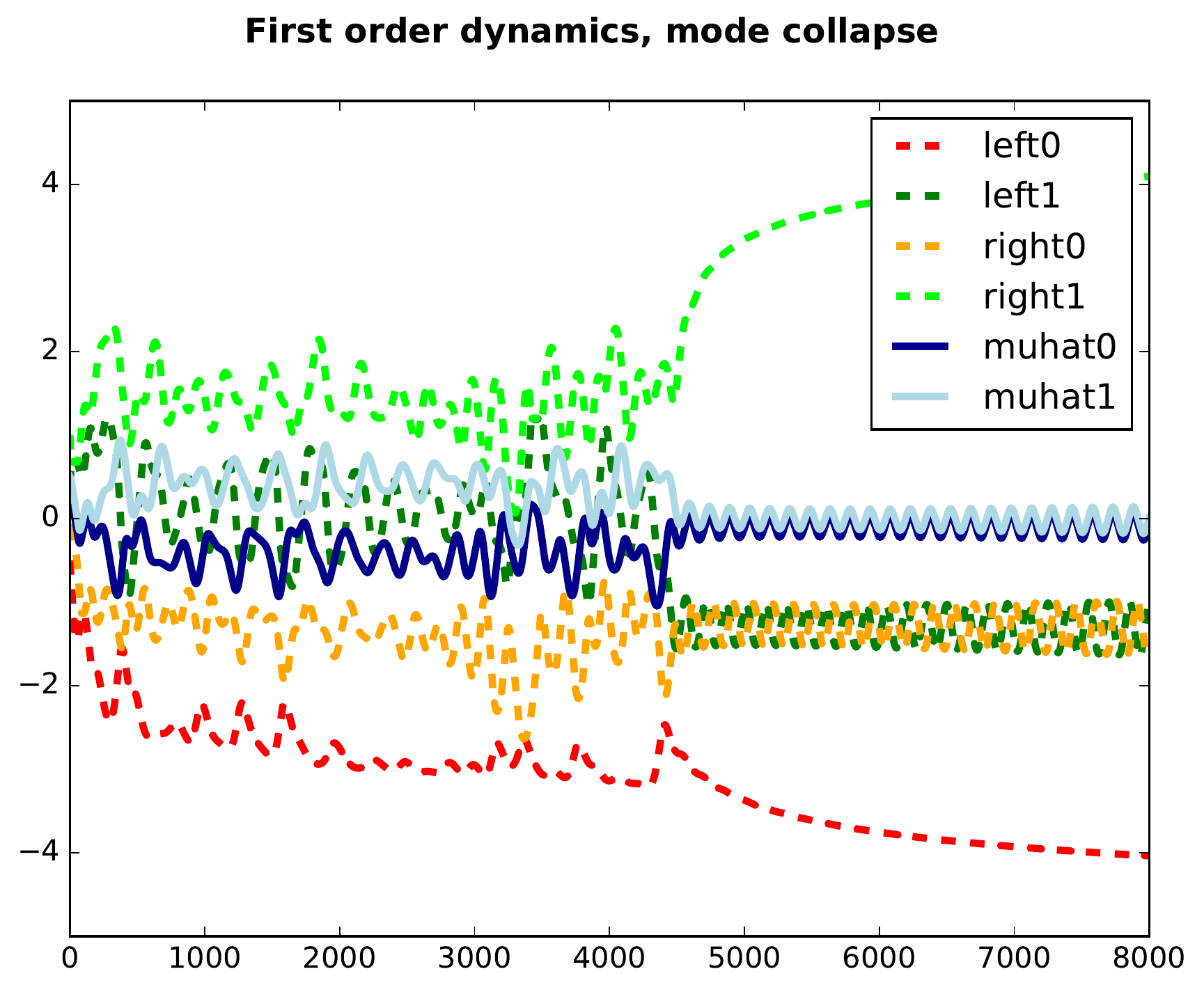} & \includegraphics[width=0.41\textwidth,clip,trim=0cm 0.5cm 0cm 0cm]{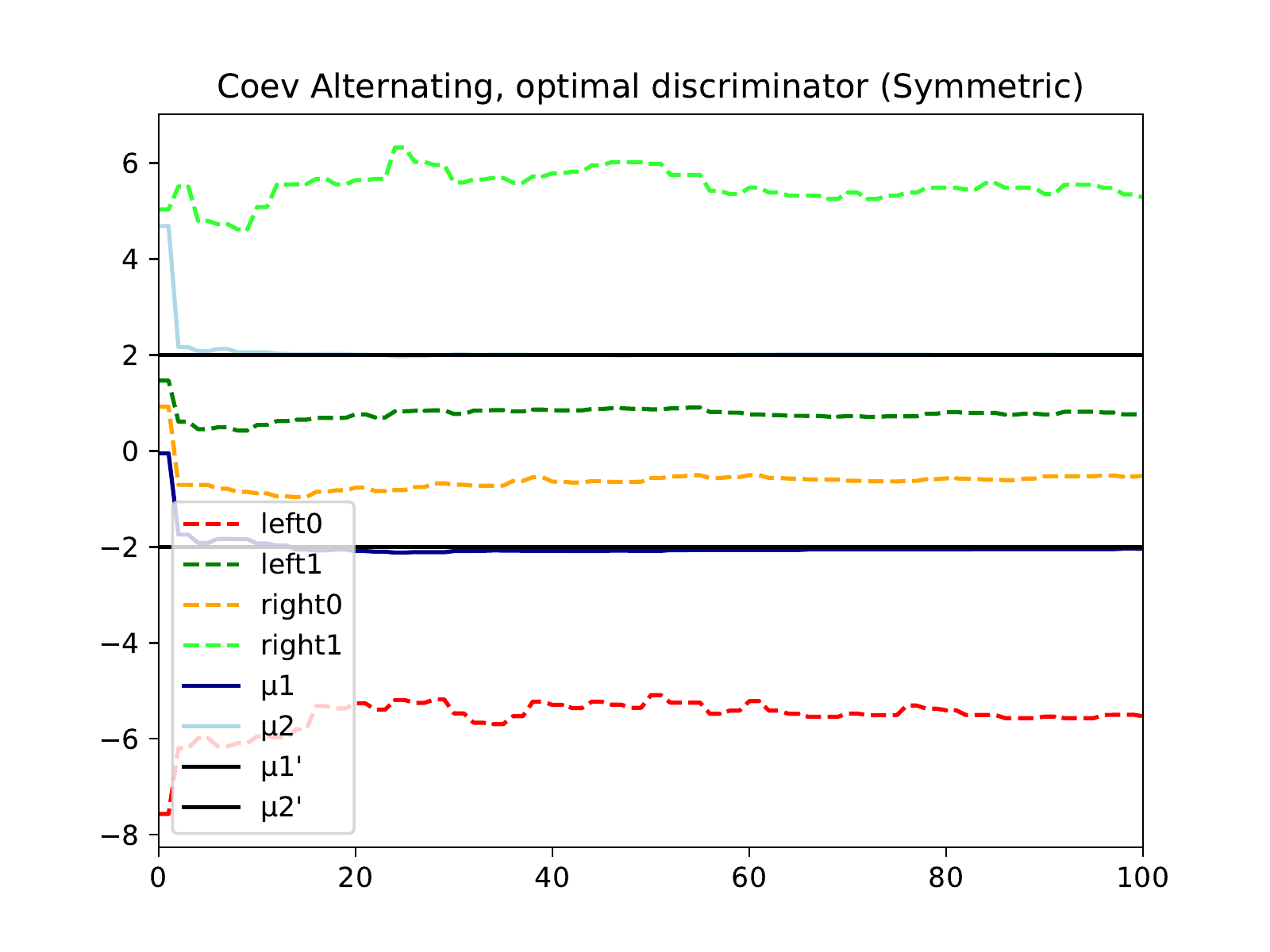} \\
			\includegraphics[width=0.35\textwidth]{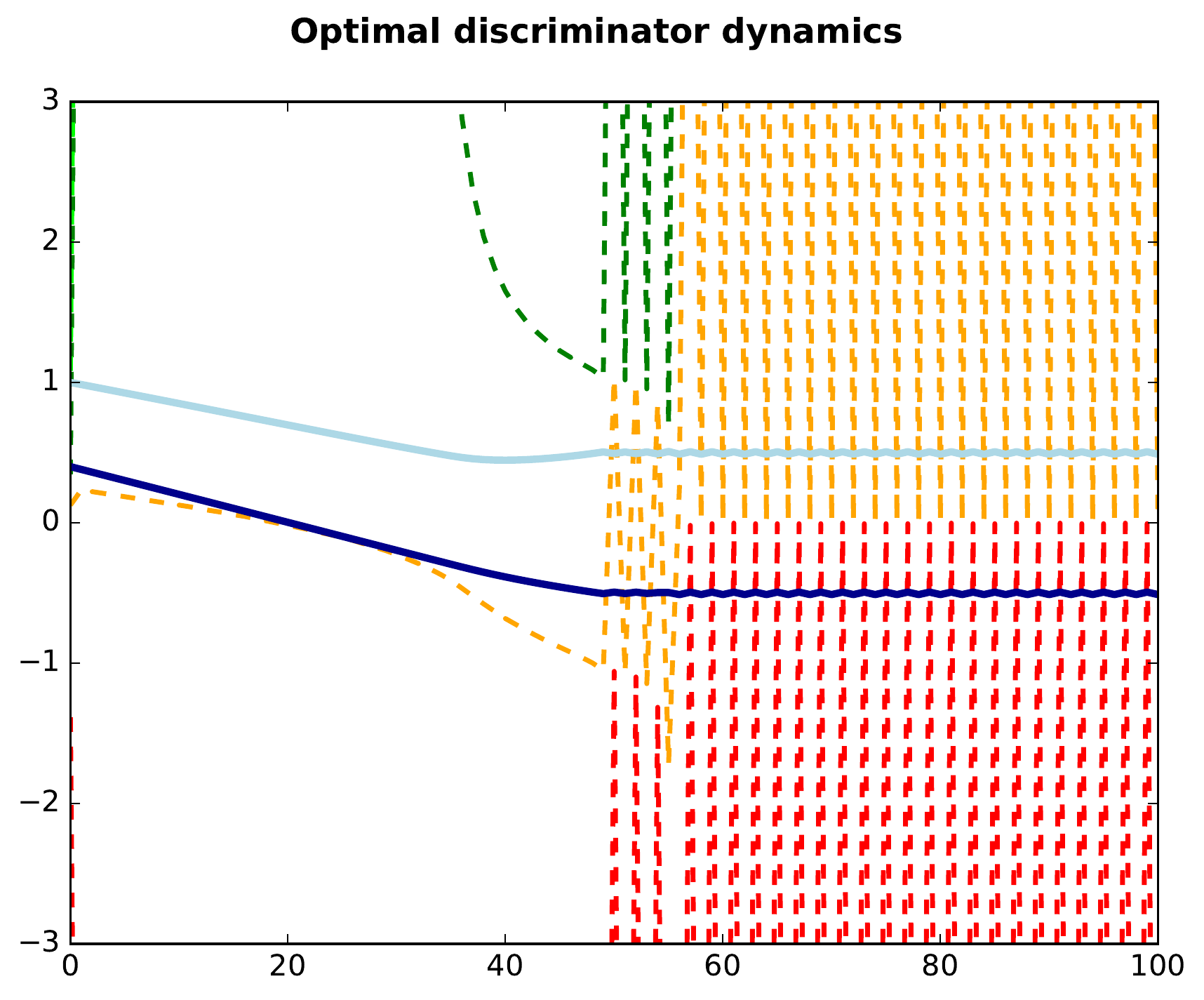}  & 
			\includegraphics[width=0.41\textwidth,clip,trim=0cm 0.5cm 0cm 0cm]{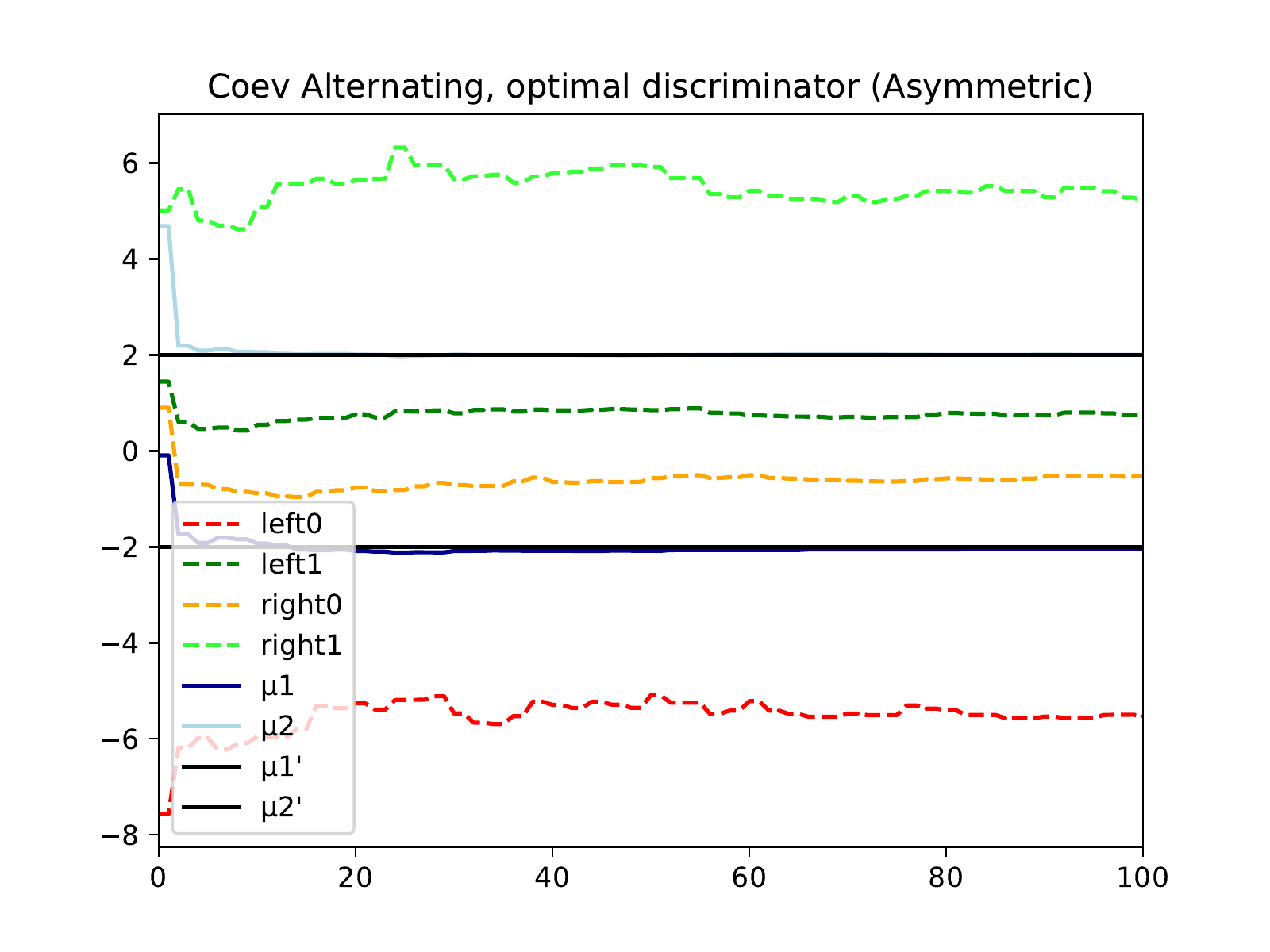}\\
			\includegraphics[width=0.35\textwidth]{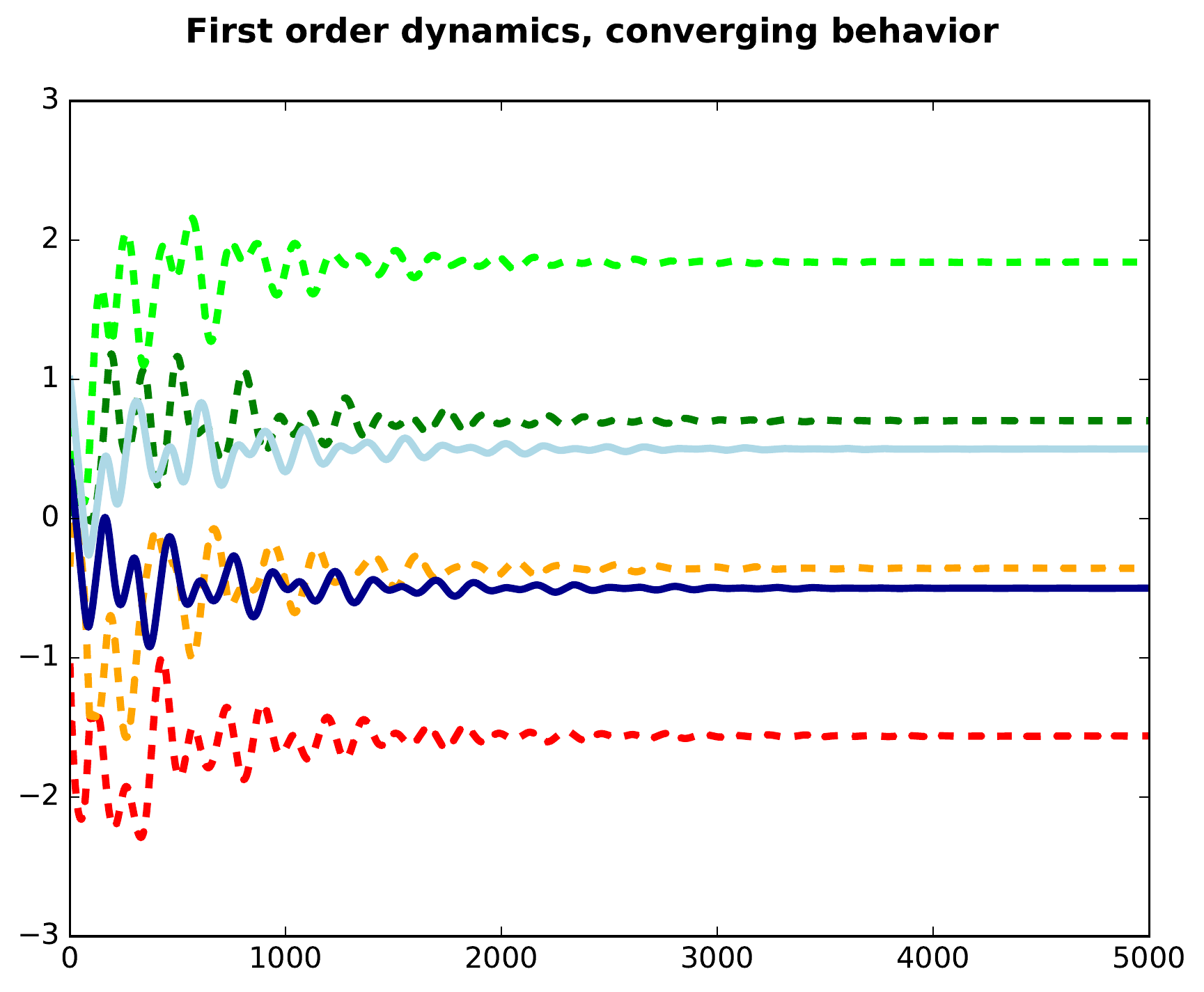}  & 
			\includegraphics[width=0.41\textwidth,clip,trim=0cm 0.5cm 0cm 0cm]{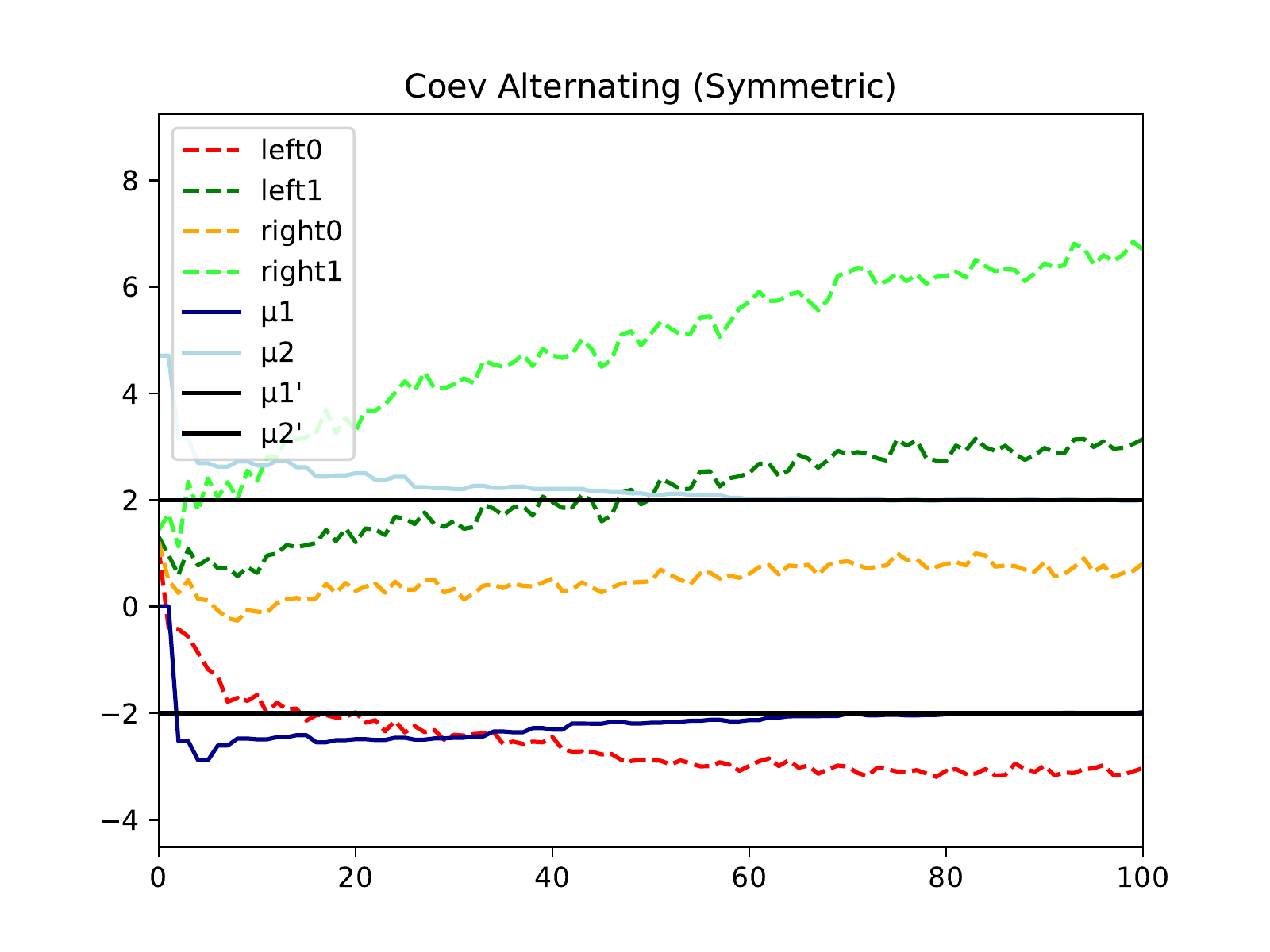}\\
			\includegraphics[width=0.35\textwidth]{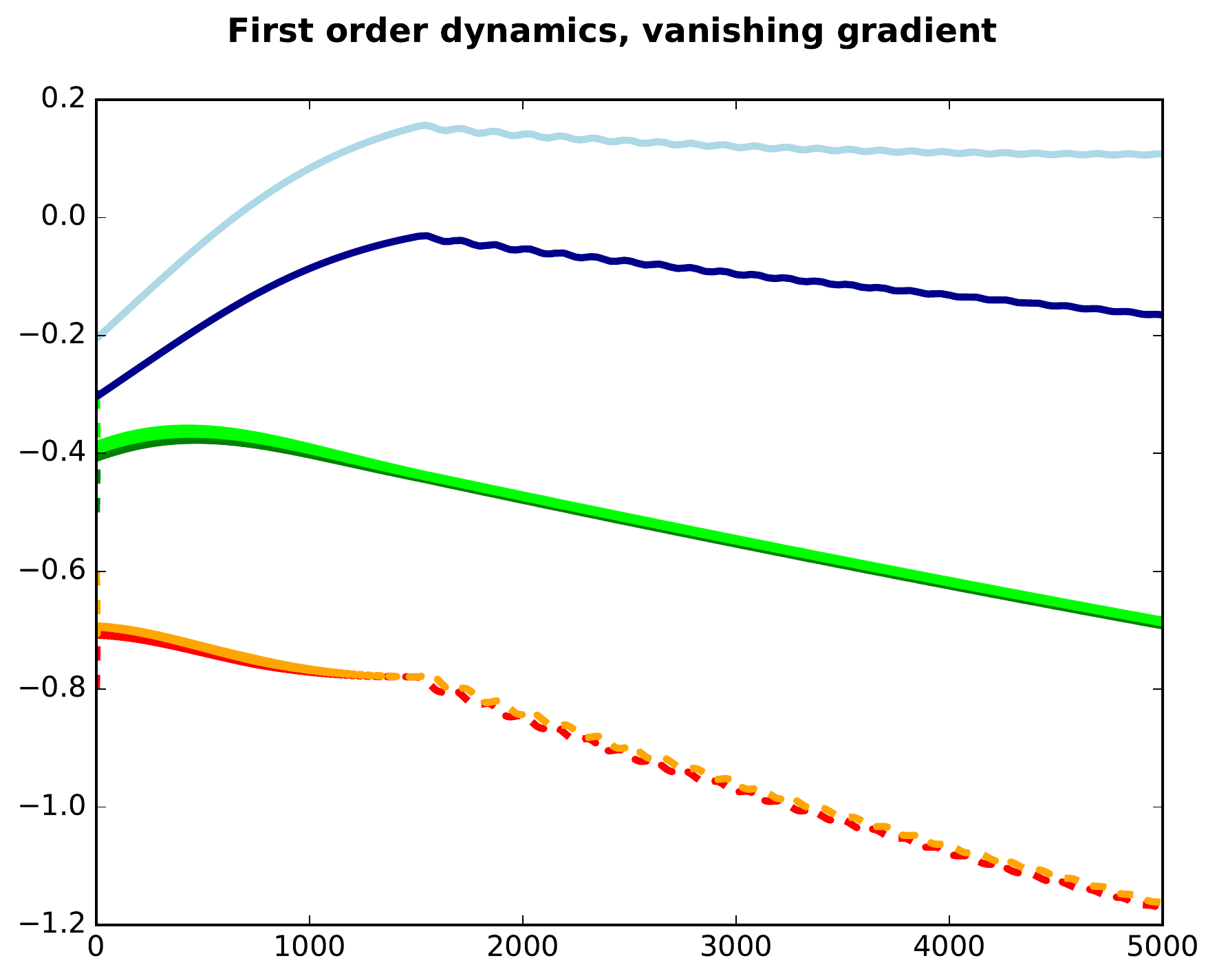}  & 
			\includegraphics[width=0.41\textwidth,clip,trim=0cm 0.5cm 0cm 0cm]{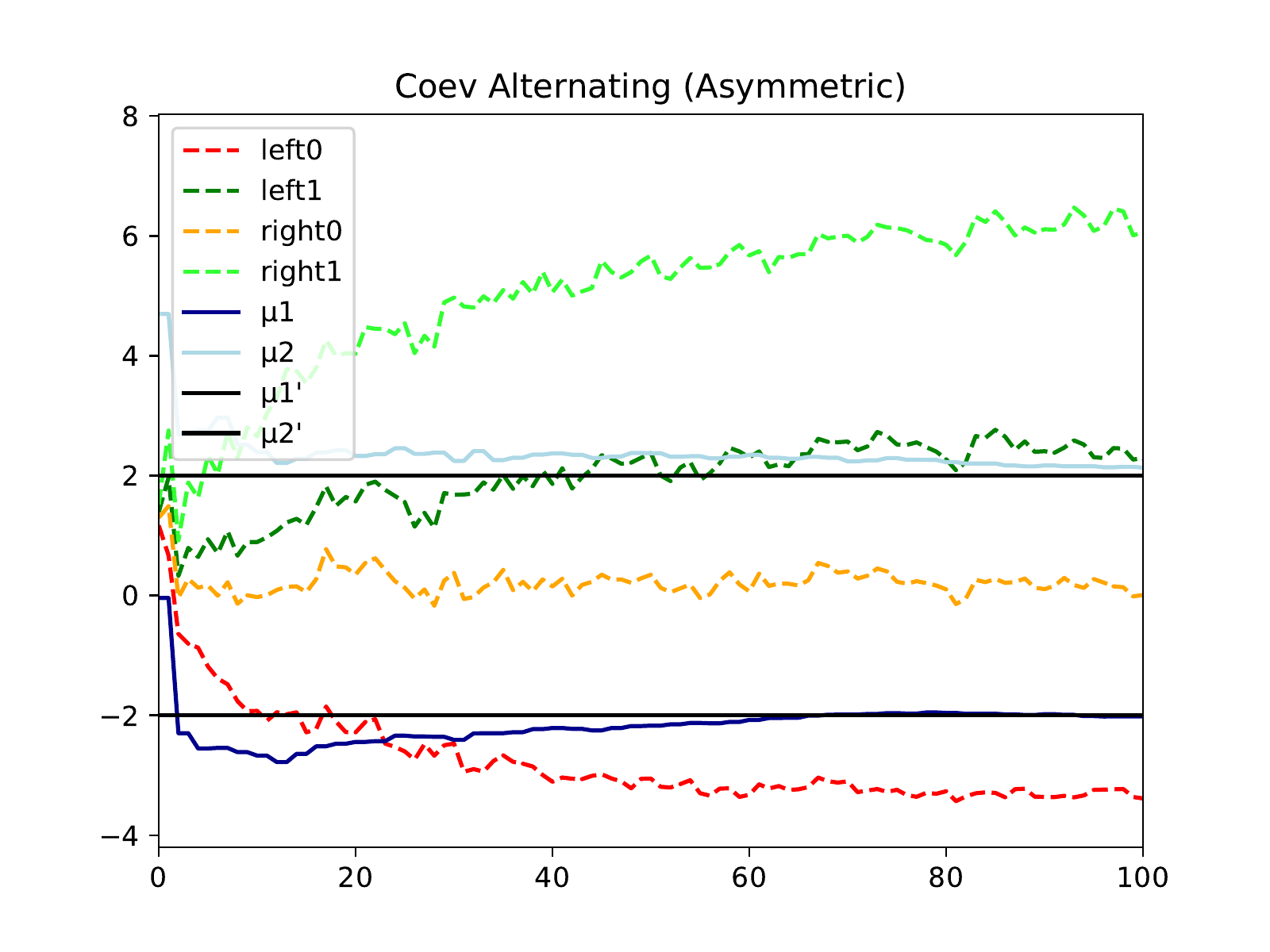}\\
			(a) & (b) \\
		\end{tabular}	
	}
	\caption{\small   Parameters convergence for the theoretical GAN model with (a) gradient-based~\cite{li2017towards} and (b) coevolutionary-based (Algorithm~\ref{alg:basic-coev-gan}) dynamics; the curves trace the best individuals (i.e., $u_1$ and $v_1$) of each generation.}
	\label{fig:gan-convergence}
\end{figure}

 \emph{Discriminator collapse.}  This term describes a phenomenon where the discriminator is stuck in a local minimum \cite{li2017towards}. Due to their local nature of updates, gradient-based dynamics are generally not able to escape these local minima without further enhancements---a problem that global optimizers like evolutionary algorithms handle better. Our results in Fig.~\ref{fig:discriminator-collapse-progress}~(a) support this proposition, using the same setup as in the previously described experiment. In particular, note the high success probability for the bottom left quadrant, where both bounds of the discriminator lie where the fitness value (Eq.~\ref{eq:loss}) is less than 0. Fig.~\ref{fig:discriminator-collapse-progress}~(b) shows an example of such bounds. In this setup, gradient-based dynamics force the bounds to collapse (i.e., $\ell_1=r_1$, $\ell_2=r_2$, see \cite[Fig. 2 (c)]{li2017towards}). On the other hand,  coevolution is able to step out of the local minimum and converges to near-optimality as shown in Fig.~\ref{fig:discriminator-collapse-progress}~(c)---with more generations, the left bound asymptotically moves towards $-\infty$. For this scenario, the parameters of $G$ were fixed to $\mu_1=-1, \mu_2=2.5$ during the whole evolutionary process.

\subsection{GAN for Images}

\paragraph{Setup.}
If not stated otherwise, the experiments were conducted with Algorithm~\ref{alg:adv-coev-gan} on a grid size of 2x2, and a population size of one per cell (i.e. one generator and one discriminator); despite this small size, the shown results are already promising in solving the pathologies described above. We leave experiments with larger grid size for future work and upcoming versions of this paper. At the end of each generation, the current cells individual is replaced with the highest ranked offspring individual created from the neighborhood. For gradient-based mutations of the neural net parameters, we use the Adam optimizer \cite{kingma2014adam} with an initial learning rate of 0.0002, which is altered with a mutation space of $\mathcal{N}(0, 1e^{-7})$ per generation. The mixture weights are updated by an (1 + 1) ES, with the mutation space of $\mathcal{N}(0, 0.01)$. Regarding the neural network topology, we used a four-layer perceptron with $700$ neurons for MNIST~\cite{lecun1998mnist}, and the more complex deconvolutional GAN architecture~\cite{radford2015unsupervised} for the CelebA~\cite{liu2015faceattributes} dataset. We use the \emph{classic} GAN setup~\cite{goodfellow2014generative} instead of recent propositions (e.g.,  WGAN~\cite{arjovsky2017wasserstein}). This simplifies the observation of interesting pathologies, which can be more complicated to precipitate with stable GAN implementations.

\begin{figure}[t]
	\centering
	\resizebox{0.5\textwidth}{!}{
	\begin{tabular}{ccc}
		\includegraphics[width=0.14\textwidth]{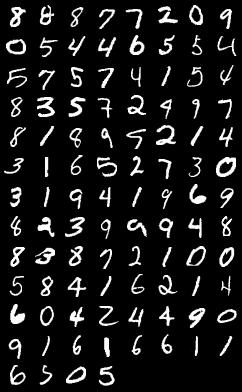} & 
		\includegraphics[width=0.14\textwidth]{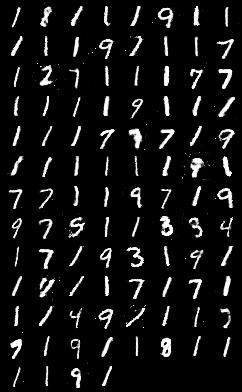} & 
		\includegraphics[width=0.14\textwidth]{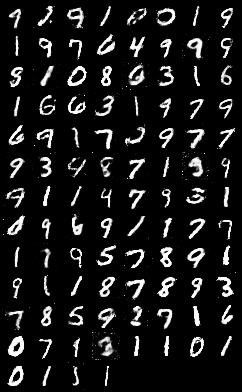} \\
		(a) Source data & (b) Mode collapse & (c) \horse
	\end{tabular}
}
	\caption{\small  Results on the MNIST dataset. (a) contains samples from the original dataset, while (b) shows a typical example for mode collapse; the generator is primarily focused on the characters 1, 9 and 7. The data sampled from a generator trained with \horse in (c) shows that coevolution is able to create higher diversity among the covered modes.}
	\label{fig:mnist}
\end{figure}

\begin{figure}[t]
	\centering
	\resizebox{0.5\textwidth}{!}{
	\begin{tabular}{cccc}
		\includegraphics[width=0.24\textwidth]{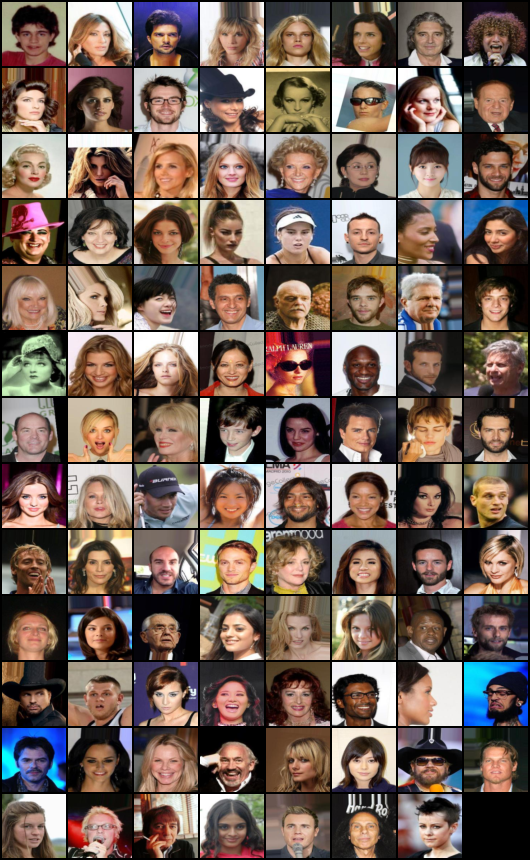} &
		\includegraphics[width=0.24\textwidth]{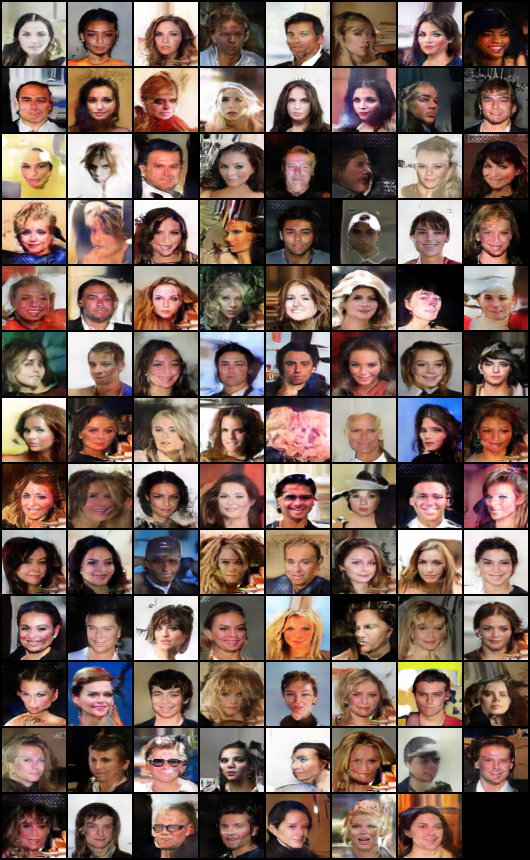} &
		\includegraphics[width=0.24\textwidth]{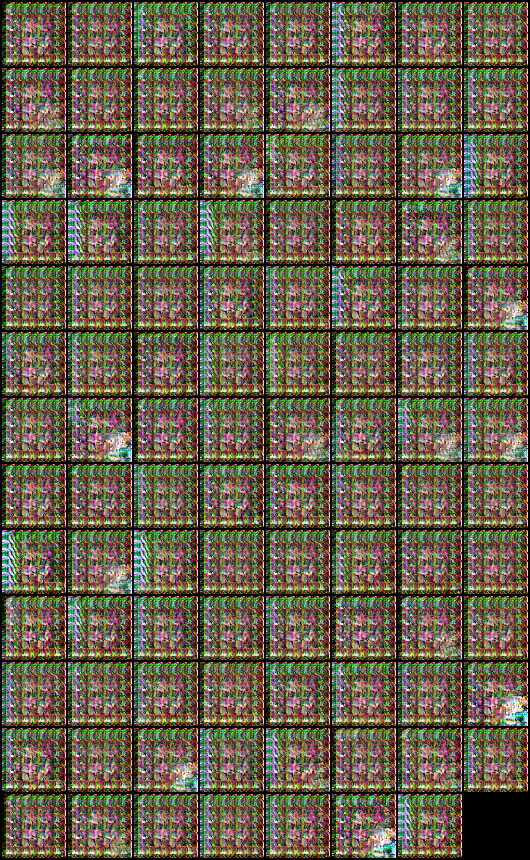} &
		\includegraphics[width=0.24\textwidth]{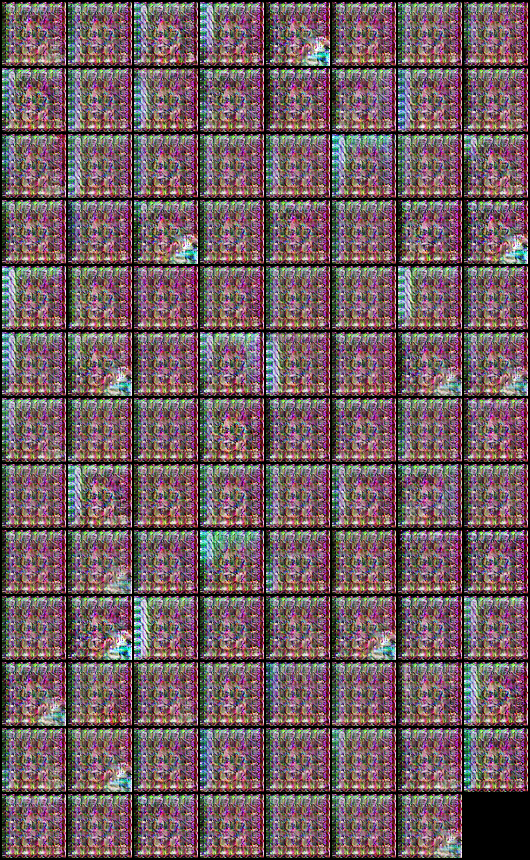}\\
		(a) Source data & (b) Before collapse & (c) First collapsed generation & (d) 10 generations after collapse
	\end{tabular}
}	
	\caption{\small   Sequence of images generated without \horse based on the CelebA dataset (b) before, (c) during, and (d) 10 generations after the systems mode or discriminator collapses. The original images in (a) are shown for comparison. This figure illustrates that, without further optimizations, DCGAN is mostly not able to step out of this scenario.}
	\label{fig:celeba}
\end{figure}

\begin{figure}[t]
	\centering
	\resizebox{0.5\textwidth}{!}{
	\begin{tabular}{cccc}
		\includegraphics[width=0.24\textwidth]{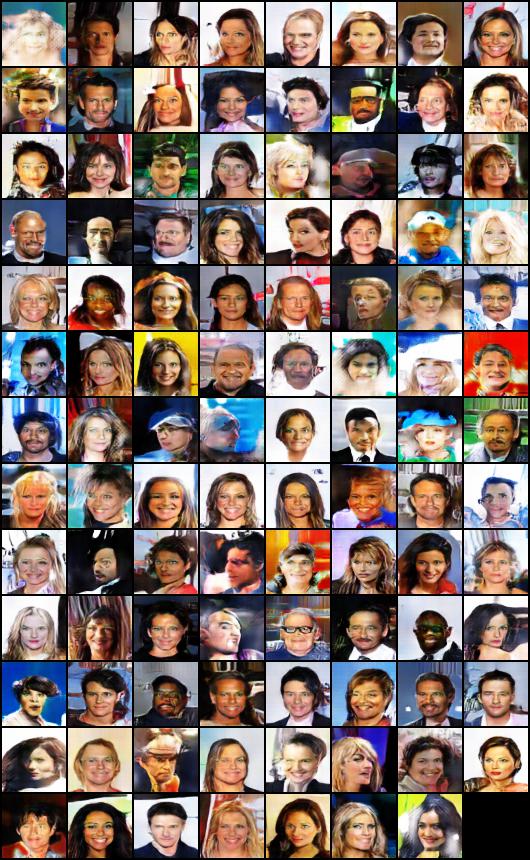} &
		\includegraphics[width=0.24\textwidth]{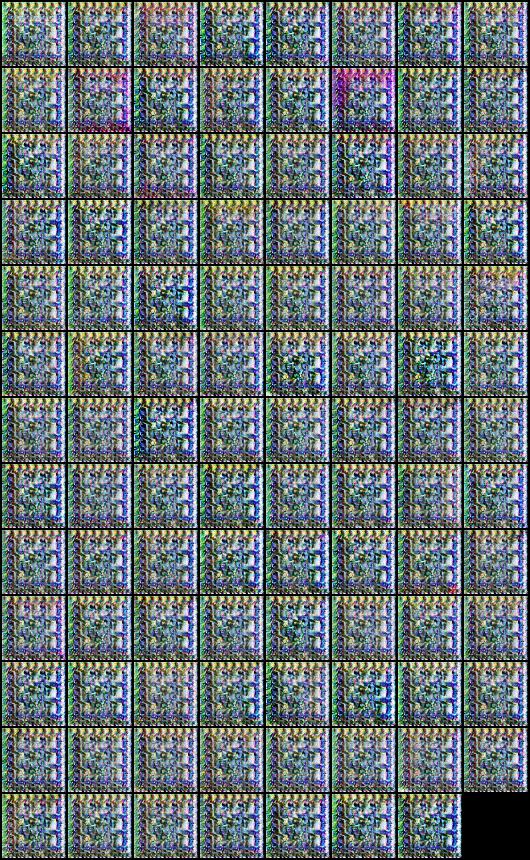} &
		\includegraphics[width=0.24\textwidth]{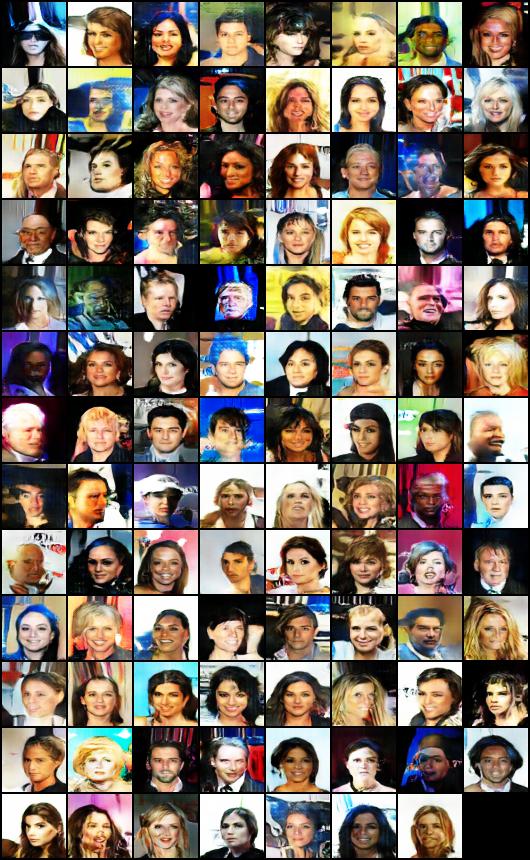} &
		\includegraphics[width=0.24\textwidth]{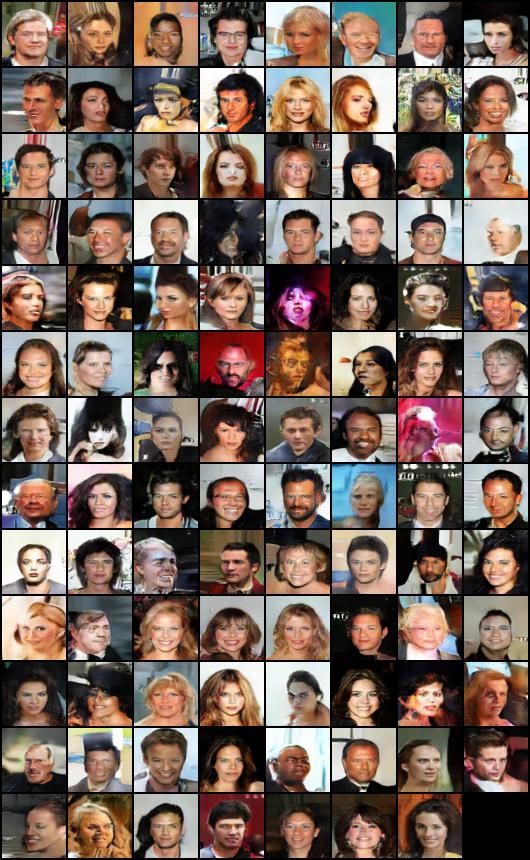} \\
		(a) Before collapse & (b) Collapsed generation & (c) One generation after collapse & (d) After 30 generations
	\end{tabular}}
	\caption{\small   Sequence of CelebA images generated by \horse (a) before, (b) during, and (c) one generation after the systems mode or discriminator collapses. Additionally, (d) shows the results generated after 30 generations. Especially when compared to Figure~\ref{fig:celeba}, this illustrates how \horse is able to overcome collapsed GANs.}
	\label{fig:celeba-lpz}
\end{figure}

\paragraph{Results.}
As stated, we conducted our experiments on two different datasets, which were selected because of their ability to show the behaviors we are primarily interested in. The MNIST dataset \cite{lecun1998mnist} has been widely used in research, and especially appropriate for showing mode collapse due to its limited target space (namely the characters 0-9). Fig.~\ref{fig:mnist} illustrates this behavior, and how \horse is able to prevent collapsing on few specific modes (numbers). Both results were generated after $400$ generations of training on the MNIST dataset with the above-mentioned four-layer perceptron. We furthermore show comparable promising results for the CelebA dataset \cite{liu2015faceattributes}, which contains more than $200,000$ images of over $10,000$ celebrities' faces. Fig.~\ref{fig:celeba} shows that a non-coevolutionary DCGAN \cite{radford2015unsupervised} collapses at a certain point and is unable to recover even after 10 more generations (with each generation processing the whole dataset). Fig.~\ref{fig:celeba-lpz} shows that the same GAN wrapped in the \horse framework is able to step out of the collapse in only the next generation. We furthermore note that, while the DCGAN collapse is easily repeatable, \horse was able to completely avoid this scenario in most of our experiments.

\section{Conclusion}
\label{sec:conclusion}
In this paper, we have investigated coevolutionary (in particular competitive) algorithms as an option to enhance the performance of gradient-based GAN training methods. We presented \horse, a framework that combines the advantages of gradient-based optimization for GANs with those of coevolutionary systems, and allows scaling over a distributed spatial grid topology. As demonstrated, our framework shows promising results on the conducted experiments, even without scaling to larger dimensions than other comparable approaches~\cite{arora2017generalization} do. Even better results may be achieved by including improved GAN types like the recently introduced WGAN \cite{arjovsky2017wasserstein}. 

\bibliographystyle{plain}
{
	\tiny
	\setlength{\bibsep}{0.0pt}
\bibliography{bibliography}

}
\end{document}